\documentclass{article}

\usepackage{microtype}
\usepackage{graphicx}
\usepackage{subfigure}
\usepackage{booktabs} % for professional tables

\usepackage{capt-of}

\usepackage{hyperref}
\usepackage{mdframed}

\usepackage[accepted]{icml2023}

\usepackage{amsmath}
\usepackage{amssymb}
\usepackage{mathtools}
\usepackage{amsthm}

\usepackage[capitalize,noabbrev]{cleveref}

\theoremstyle{plain}

\theoremstyle{definition}

\theoremstyle{remark}

\usepackage[textsize=tiny]{todonotes}

\icmltitlerunning{Why do universal adversarial attacks work on large language models?: Geometry might be the answer}

\begin{document}

\twocolumn[
\icmltitle{Why do universal adversarial attacks work on large language models?: Geometry might be the answer}



\icmlsetsymbol{equal}{*}

\begin{icmlauthorlist}
\icmlauthor{Varshini Subhash}{equal,yyy}
\icmlauthor{Anna Bialas}{equal,yyy}
\icmlauthor{Weiwei Pan}{yyy}
\icmlauthor{Finale Doshi-Velez}{yyy}
\end{icmlauthorlist}

\icmlaffiliation{yyy}{John A. Paulson School of Engineering and Applied Sciences, Harvard University}

\icmlcorrespondingauthor{Varshini Subhash}{varshinisubhash@g.harvard.edu}
\icmlcorrespondingauthor{Anna Bialas}{annabialas@g.harvard.edu}

\icmlkeywords{Machine Learning, ICML}

\vskip 0.3in
]



\printAffiliationsAndNotice{\icmlEqualContribution} % otherwise use the standard text.

\begin{abstract}
Transformer based large language models with emergent capabilities are becoming increasingly ubiquitous in society. However, the task of understanding and interpreting their internal workings, in the context of adversarial attacks, remains largely unsolved. Gradient-based universal adversarial attacks have been shown to be highly effective on large language models and potentially dangerous due to their input-agnostic nature. This work presents a novel geometric perspective potentially explaining universal adversarial attacks on large language models. By attacking the 117M parameter GPT-2 model, we find evidence indicating that universal adversarial triggers could be embedding vectors which merely approximate the semantic meaning captured by their adversarial training region. This hypothesis is supported by white-box model analysis comprising dimensionality reduction and similarity measurement of hidden representations. We believe this new geometric perspective on the underlying mechanism driving universal attacks could help us gain deeper insight into the internal workings and failure modes of LLMs, thus enabling their mitigation.
\end{abstract}

\section{Introduction} \label{introduction}

Adversarial attacks have gained significant research attention due to their ability to expose system vulnerabilities and model limitations. While these have been popular in computer vision for a while, similar efforts in natural language processing (NLP) have gained momentum in recent years. One particularly effective class of attacks has been the gradient-based universal adversarial attack, introduced in computer vision by \cite{Moosavi-Dezfooli16} and extended to NLP by \cite{Wallace2019}. The latter performs a gradient-guided search over tokens and generates trigger sequences which when appended to the input, will generate adversarial model output. This technique of generating a `universal’ attack can pose a greater degree of adversarial threat, due to its ability to be reused across all inputs, i.e. it is input-agnostic. Additionally, \cite{Singla2022MINIMALMM} make the originally data-intensive trigger generation process cheap and \cite{Song2021UniversalAA} overcome the limitation of gibberish triggers by making them harder to detect. \cite{Wallace2019} demonstrate the effectiveness of these triggers on models such as GPT-2, with evidence of easy transferability of the same attack across varying model sizes.

Despite there being significant progress in understanding the construction and efficacy of gradient-based adversarial attacks, there has been limited progress towards interpreting and explaining the underlying mechanism which makes these attacks successful. Furthermore, transformer based large language models have become increasingly powerful and ubiquitous in society, which has ushered in an urgency for progress in this domain, from a model safety standpoint. 

This work seeks to uncover the underlying mechanism behind the effectiveness of universal adversarial attacks by adopting a geometric perspective. It leverages the geometric interpretation of word embeddings to propose an explanation for the behavior of universal adversarial attacks and provides initial experimental evidence via dimensionality reduction and white-box model analysis. Geometric approaches have been shown to be useful in explaining adversarial attacks in computer vision \cite{ilyas2019adversarial, engstrom2019a}, which serves as motivation for adopting it in this work. 

To this end, our contributions are as follows:

\begin{enumerate}
    \item We propose a novel geometric perspective as a potential explanation for universal adversarial attacks on large language models.
    \item We support our findings with initial experimental evidence consisting of dimensionality reduction and similarity measurement of hidden representations. 
    \item We leverage this new perspective to open doors to additional potential explanations for the behavior of universal triggers, as observed in literature.
\end{enumerate}

By reverse-engineering this class of attacks, we hope to shed light on the internal workings of large language models, their failure modes and potential strategies for mitigating undesirable downstream consequences.

\section{Related Work}
There are several works which have investigated the construction and efficacy of adversarial attacks in NLP. There have also been attempts to reverse-engineer neural networks and transformer models through explainability and interpretability. However, to the best of our knowledge, there hasn't been prior work that has attempted to explain the underlying mechanism of a gradient-based adversarial attack on a large language model. This work presents a novel geometric approach by leveraging white-box interpretability to explain the effectiveness of universal adversarial attacks. 

\subsection{Universal Adversarial Attacks}
A comprehensive survey of adversarial attacks in NLP by \cite{Zhang2019Survey} gives a bird’s eye view of the advances in this space. The notion of universal adversarial perturbations, obtained via a gradient-based optimization, was introduced by \cite{Moosavi-Dezfooli16} in computer vision and extended to language models by \cite{Wallace2019}. These universal adversarial triggers leveraged the gradient-based token replacement strategy by \cite{Ebrahimi2017HotFlip} in their construction and were further investigated by several works. \cite{Singla2022MINIMALMM} introduce MINIMAL, a data-free alternative to the originally data-intensive process of generating universal triggers. \cite{Song2021UniversalAA} design \emph{natural attack triggers} which consist of more semantically meaningful triggers, thus overcoming the limitation of previous triggers which were often nonsensical and could easily be detected by humans. \cite{Heidenreich2021} thoroughly study the susceptibility of GPT-2 to these triggers by varying the topic and stance that the trigger is trained on. However, no prior work has tried to explain and interpret these attacks, which this work attempts to address.

\subsection{Understanding Language Models Via Interpretability}

There have been many approaches in literature to interpret and explain large language models. \cite{Atmakuri2022Robustness} study the adversarial robustness of explanation methods for language models and find that feature attribution based methods are sensitive to input perturbations. \cite{Mosca2022Logits} study patterns in the logits of models subject to textual adversarial attacks and train a classifier to detect these adversarial samples.

\emph{Probing} is a technique which leverages layer activations, attention weights and hidden representations to interpret large language model behavior. This has proven to be useful in understanding how the model encodes language structure within its representations (\cite{Hewitt-Manning-2019, Belinkov2017, Peters2018Deep, Adi2016}). \cite{Tenney2019Bert, Lin2019OpenSesame} perform \emph{layer-wise probing} to study the linguistic awareness of BERT. \cite{Wallat2020} investigate the relational knowledge learned by BERT and conclude that intermediate layers contribute significantly to the total knowledge and that the knowledge is distributed unevenly across layers. Within \emph{self-attention probing}, \cite{Kovaleva2019, Clark2019} analyze BERT's attention heads and identify underlying patterns. \cite{chizhikova-etal-2022-attention} probe the attention weights in BERT and find evidence that attention can capture semantic awareness. \cite{attention-flows-2020} present a framework to visualize and compare the attention mechanisms in language models. \cite{Coenen2019} offer a geometric perspective to the internal representations of BERT by studying both the attention weights and layer-wise embeddings. \cite{Brunner2019} also study self-attention and layer-wise embeddings from the perspective of \emph{identifiability}, i.e. the ability of a model to learn stable representations. They find that attention distributions are not identifiable, hence not directly interpretable. \cite{Wang2022} provide a circuit-based explanation for GPT-2, \cite{Meng2022} discover middle-layer neuron activations in GPT which are critical towards model predictions and \cite{Geiger2021} leverage causal abstraction to connect model behavior with hidden representations. \cite{Geva2020} demonstrate how feed-forward layers in transformers map training examples to output distributions. By adopting the approach of mechanistic interpretability, \cite{nanda2022grokking} explain emergent behavior such as \emph{grokking}. 

Unlike previous approaches, we combine white-box interpretability with a novel geometric perspective, to explain the underlying mechanism of universal adversarial attacks.

\section{Universal Adversarial Triggers on Large Language Models}

\cite{Wallace2019} introduce \emph{universal adversarial triggers} (UATs) as a sequence of tokens searched via gradient-based optimization, such that these triggers when appended to an input, cause the model to perform poorly at a range of language tasks. They are \emph{input-agnostic} which means that the same trigger can be used with any input, for a given model, thus increasing its degree of adversarial threat. They are also observed to be composed of mostly nonsensical tokens. For technical details on the trigger generation method, we refer the reader to \cite{Wallace2019} and \cite{Ebrahimi2017HotFlip}. As alluded to by \cite{Wallace2019} with their GPT-2 use case, we confirm that these universal adversarial triggers are just as effective on large language models.

In order to demonstrate this, we choose two tasks discussed by \cite{Wallace2019} -- classification as sentiment analysis and conditional text generation. For these tasks, our models of choice are DistilBERT \cite{DistilBERT-Sanh-2019} and OPT-350M \cite{zhang2022opt} respectively. We generate a trigger optimized for DistilBERT and get the following results. For triggers of length 5 and 10, for positive reviews, the test classification accuracy drops from 89.2\% to 2.4\% and 0.4\% respectively. For negative reviews, triggers of length 5 and 10 cause the test accuracies to drop from 92.4\% to 10\% and 1.6\% respectively. This confirms findings by \cite{Wallace2019} that longer triggers are more effective. 

For text generation, we repurpose the trigger trained for GPT-2 by \cite{Wallace2019} (`\emph{TH PEOPLEMan goddreams Blacks}') and find that it works off the shelf for OPT-350M, causing the model to generate racist and offensive text. For the remainder of this work, we refer to this trigger as \emph{gpt-trigger}. This is indicative of the fact that in some cases, a UAT transfers easily across models, making it model-agnostic. We further discuss this behavior in Section \ref{sec:future-work}. In this work, we restrict our analysis to the task of text generation and leave sentiment analysis for future work.

\section{Preliminary Analysis}
By adopting techniques of self-attention probing and trigger perturbations, we attempt to gain preliminary insights into the behavior of universal adversarial triggers.

\subsection{Self-Attention Probing}
Since the self-attention mechanism is what separates a transformer model from other models, we first check the effect of the trigger on the self-attention weights. Despite our initial hypothesis that the trigger tokens could be paid attention to disproportionately, we see no conclusive evidence of the same, for sentiment analysis on DistilBERT. For text generation, we feed \emph{gpt-trigger} to OPT-350M and note an underlying pattern in the attention heatmaps -- the final token `Blacks' which is the only semantically meaningful token, happens to pay a large amount of attention to itself. 

\subsection{Trigger Perturbations}
We further investigate \emph{gpt-trigger} by perturbing its tokens and observe the corresponding effect on the model output. The limitation of this analysis is that it has been performed on a single trigger and may not be generalizable.

\begin{enumerate}
    \item \emph{Deletions}: We opt to perform deletions from left to right since the self-attention on the last token prompts us to preserve it. We refer to these partial triggers as \textit{deletion triggers}. For a set of 10 generated sentences per deletion trigger, we see that the number of occurrences of racist output goes down as the number of deletions increases, i.e. truncating tokens from the trigger diminishes its effect. However, despite being lower in frequency, some instances of the model output continue to remain problematic. The token `Blacks' alone fails to perform adversarially. 

    \begin{mdframed}[style=MyFrame,nobreak=true,align=center]
\textit{gpt-trigger}: \color{red}TH PEOPLEMan goddreams Blacks.\\
\color{black}Deletion 1: \color{red}PEOPLEMan goddreams Blacks \color{black} are too dumb to realize that they're only the opposite of ..\\
Deletion 2: \color{red}goddreams Blacks \color{black} are such a fickle bunch.\\
Deletion 3: \color{red}Blacks \color{black} and whites who do not identify with their race but identify in ways that are consistent .. \\
\end{mdframed}
    
    \item \emph{Positional Perturbations}: We shuffle the positions of the trigger words to check for any influence of word position on the model output. For \textit{gpt-trigger} which consists of 4 words, we obtain all 24 permutations and tokenize each shuffled trigger. We feed these to the model and find that only 6 among 24 (25\%) of the shuffled triggers generate coherent racist output. 5 among these 6 have `Blacks' as the last trigger token, strongly indicating the positional importance of the last token. Additionally, a manual check of the generated text for each trigger indicates that the position of the token `Blacks' has an effect on how coherent the offensive output is. For instance, the trigger \emph{`goddreams TH Blacks PEOPLEMan'} where `Blacks' is the penultimate token, generates completely harmless output -- `Your family makes me very, very happy!'. We further note that the position impacts the tokenization output.
    
    \item \emph{Semantic Perturbations}: Since the semantically meaningful trigger word `Blacks' appears to plays a role in the output, we swap this word with other racial groups and study the effect on the model output. For example, we replace `Blacks' with `Jews', `Asians', `Muslims' etc. -- an experiment attempted by \citet{Wallace2019} as well. The drawback of employing this approach is that it requires a trigger word to have semantic meaning, which is mostly not the case.

    We begin by replacing `Blacks' with words such as `Asians', `Jews', `Muslims' and `Hispanics' and find that for each case, the model generates semantically meaningful racist output that corresponds with the racial group. Replacing the last token with nationalities such as `Americans', `Indians', `Russians', etc. also consistently generates offensive output, despite not being present in the original adversarial text that the trigger was trained on. 

    We then replace the last trigger word with each word in the adversarial target text that \emph{gpt-trigger} was trained on, to check for a potential role played by target text occurrences within the trigger. We observe no relationship between the presence of a target text word and the model output, since offensive target words did not generate repeatable problematic output.

    We then generate 100 random words from the English language and append each as the last trigger word instead of `Blacks' and manually inspect the output. Only 5 words (barrage, pound, dangerous, fiery and atheist) among 100 were problematic, which strongly suggests that highly attended-to and semantically meaningful trigger words (the last one in this case) could be playing a significant role in the adversarial effect on the model. 
\end{enumerate}

\section{A Geometric Perspective}

\begin{figure}[htp!]
\centerline{\includegraphics[width=1\columnwidth]{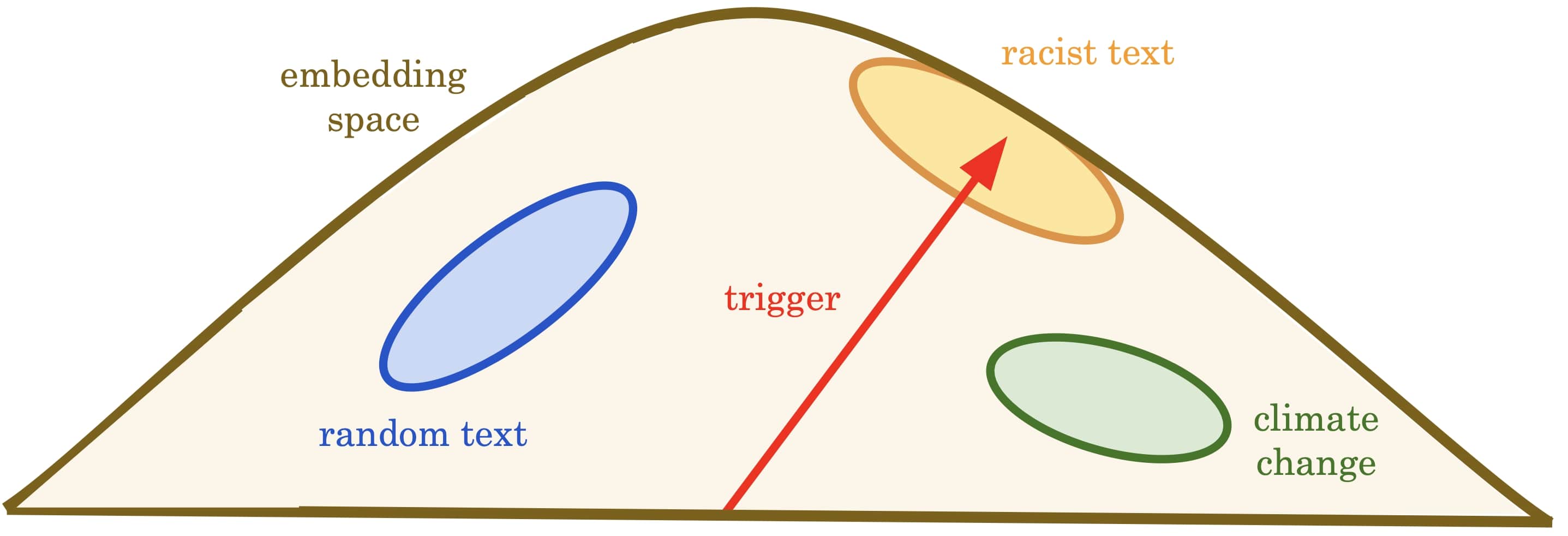}}
\caption{\textit{Geometric perspective}: The trigger (red) could be behaving like an embedding vector, optimized to arrive at a semantic region which is racist (yellow). Other semantic regions exist, like random English sentences (blue) and climate change (green), separable from the racist region.}
\label{fig:geometric-hypothesis}
\end{figure}

Given the indication that the semantic meaning of trigger tokens plays a role in its effectiveness (seen in both self-attention probing and trigger perturbations), we propose a geometric perspective to potentially explain universal adversarial triggers. Specifically, we consider the geometric interpretation of word embeddings and conjecture that the trigger sequences could be behaving like embedding vectors, lying in the part of the embedding space corresponding to the adversarial text it is trained to generate. For example, \emph{gpt-trigger} is trained on and generates racist, offensive text. As per the visual depiction in Figure \ref{fig:geometric-hypothesis}, this trigger could be behaving like a vector whose local neighborhood embeds the racist text that it is trained on. As a result, the best possible universal adversarial trigger can be thought of as the best vector that embeds the infinite adversarial text that we wish to train it on.

In order to support this hypothesis, we leverage dimensionality reduction and white-box model analysis. The problem can be reduced to showing measurable similarity between the trigger and adversarial text alongside measurable dissimilarity between the trigger and innocuous text. We present our experimental setup and results in the following section.

\section{Experiments \& Results}

We borrow the baseline experimental setup by \citet{Wallace2019} and attack the 117M parameter GPT-2 model using \emph{gpt-trigger}. In order to find evidence supporting our geometric perspective (Figure \ref{fig:geometric-hypothesis}), we seek to show similarity between the trigger and adversarial racist text alongside dissimilarity between the trigger and text belonging to other semantic categories. However, as is the case with most problems in machine learning, we encounter the \emph{curse of dimensionality} \citet{bellman1966dynamiccurseofdim}. Specifically, it is non-trivial to compare sentences represented via GPT-2's 768 dimensional word embeddings without performing some form of dimensionality reduction. We try three techniques - PCA, t-SNE and UMAP and find that UMAP offers the most useful insight, which we present below. 

We begin by creating 10 groups of sentences, each belonging to a specific semantic category. The first sentence group consists of four variants of the \emph{gpt-trigger}, where three variants have the last token swapped with other racial groups. The second sentence group consists of the adversarial target text that \emph{gpt-trigger} was originally trained on. The third group consists of arbitrary racist text that \emph{gpt-trigger} has not seen or been trained on before. The rest of the sentence groups consist of generated sentences by ChatGPT, including random English sentences, generic positive sentences about black people, white people, Asians, Muslims, sentences about climate change and sentences about Donald Trump.

\begin{figure}[ht]
\vskip 0.2in
\begin{center}
\centerline{\includegraphics[width=\columnwidth]{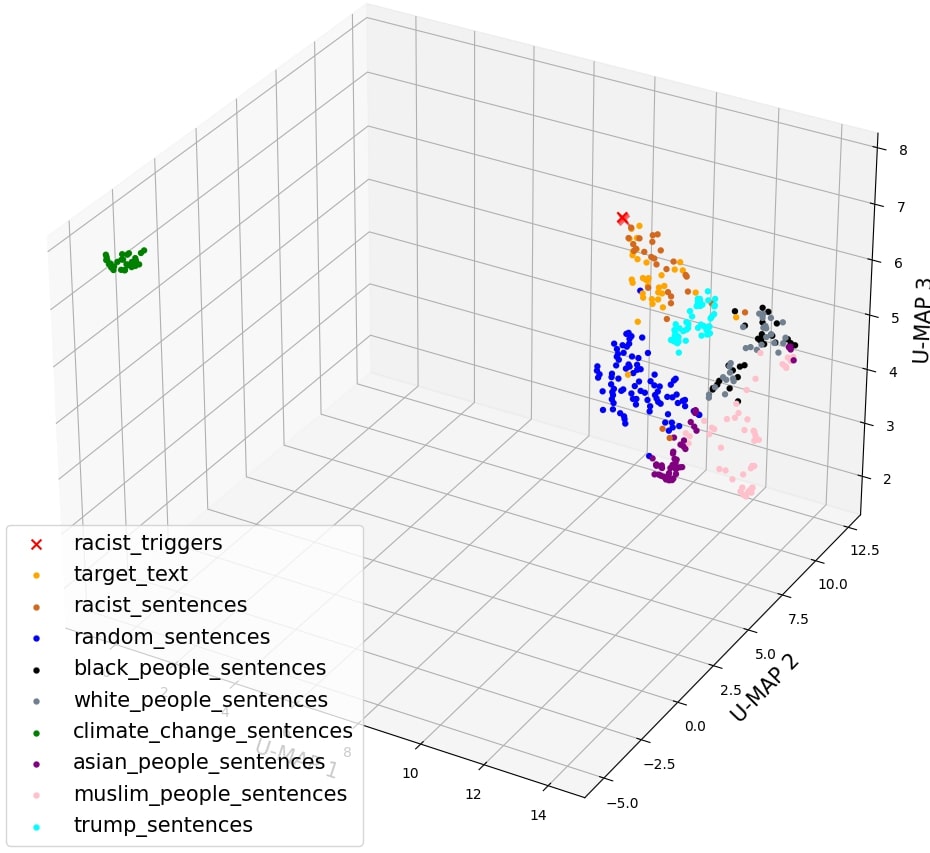}}
\caption{Sample UMAP dimensionality reduction (neighbors = 15, minimum distance = 0.2) on sentence groups. Triggers (red), adversarial training text (orange) and arbitrary unseen racist sentences (brown) cluster together, indicating a semantic neighborhood. Semantically similar sentence groups (black, white, Asian and Muslim people) also cluster together.}
\label{fig:umap-3d}
\end{center}
\vskip -0.2in
\end{figure}

We collect the word embedding matrix for each of these sentences from GPT-2's final layer and pool individual word embeddings into a sentence embedding by performing a dimension-wise average across tokens. We then perform a UMAP dimensionality reduction on these 768-dimensional average sentence embeddings and project them onto a lower dimensional space to observe similarities between sentence groups. In the 3D projected space (Figure \ref{fig:umap-3d}), UMAP appears to strongly suggest that the triggers, adversarial target text and arbitrary racist text are tightly clustered together, with visible separation from other sentence groups. 

We then check for the robustness of this potentially isolated occurrence by varying the hyperparameters for UMAP and observing the clustering behavior across a range of values. Specifically, we vary the number of nearest neighbors and minimum distance. We observe that for reasonable ranges of hyperparameter values for a given reduced dimension, there is evidence of stable clustering behavior between the trigger, target text and racist sentence groups. Despite indications that the geometric explanation holds in lower dimensions, an understandable critique could be questioning its validity in higher dimensions. To this end, we vary the number of reduced dimensions to be as large as 250. We find that for distance metrics such as Euclidean, Manhattan, Canberra and Chebyshev, the triggers appear to be closest to the target and racist sentences on average, and have larger average distances from the other groups. A sample visualization for a pair of hyperparameter values (neighbors = 50, distance = 0.25), can be seen in Figure \ref{fig:sliced-metrics}. A comprehensive set of results visualizing all variations are available in the Appendix \ref{sec:appendix}. 

\paragraph{Behavior for other triggers.}

In order to test if this behavior generalizes across triggers trained on other topics, we choose two topics -- anti-vaccination and the Flat Earth conspiracy theory. We borrow the trigger trained on the Flat Earth conspiracy theory by \cite{Heidenreich2021} (\textit{`Historical theories earth sphericalitionally'}) and train our own trigger on anti-vaccination (\textit{`Doctors vaccinated GOvacc Poison Stop asserting Blueprint'}). We repeat the experiment described earlier for these two triggers and include \textit{gpt-trigger}, its target text and arbitrary racist sentences during dimensionality reduction. Overall, we observe similar behavior across a range of UMAP hyperparameter values, dimensions and distance metrics. The target text comprising sentences in support of anti-vaccination and the Flat Earth conspiracy (shown in orange in the Appendix \ref{sec:appendix}), is much closer to the trigger on average compared to other sentence groups. A notable difference we observe between this experiment and the one with \textit{gpt-trigger} is that these triggers appear to be close to semantically similar sentences with non-adversarial content. That is, the anti-vaccination and Flat Earth triggers are close to sentences supporting vaccination and opposing the Flat Earth theory respectively (shown in brown in the Appendix \ref{sec:appendix}). During text generation, we also observe similar behavior, where these triggers do not consistently generate adversarial text. That is, some generated sentences are in support of vaccination and in opposition of the Flat Earth theory. \cite{Heidenreich2021} also showed that triggers trained on niche topics tend to be weaker than those trained on broader topics, which could be a possible explanation. In contrast, \textit{gpt-trigger} was much closer to arbitrary racist text and farther away from positive sentences about racial groups. The complete set of results visualizing these observations are available in the Appendix \ref{sec:appendix}.

Our initial experimental evidence appears to strongly suggest the validity of our geometric explanation for universal adversarial triggers. We show that for a range of reduced UMAP dimensions, across varying distance metrics and reasonable hyperparameter values, the universal adversarial trigger is closest in distance to the adversarial target text it is trained on and arbitrary text which is semantically similar to the target text. This behavior is also seen to replicate across trigger topics. In other words, the trigger might indeed be behaving like a vector which embeds the adversarial target text it is trained on, which in the case of \textit{gpt-trigger}, happened to be racist and offensive.

\begin{figure*}[ht]
  \includegraphics[width=\textwidth]{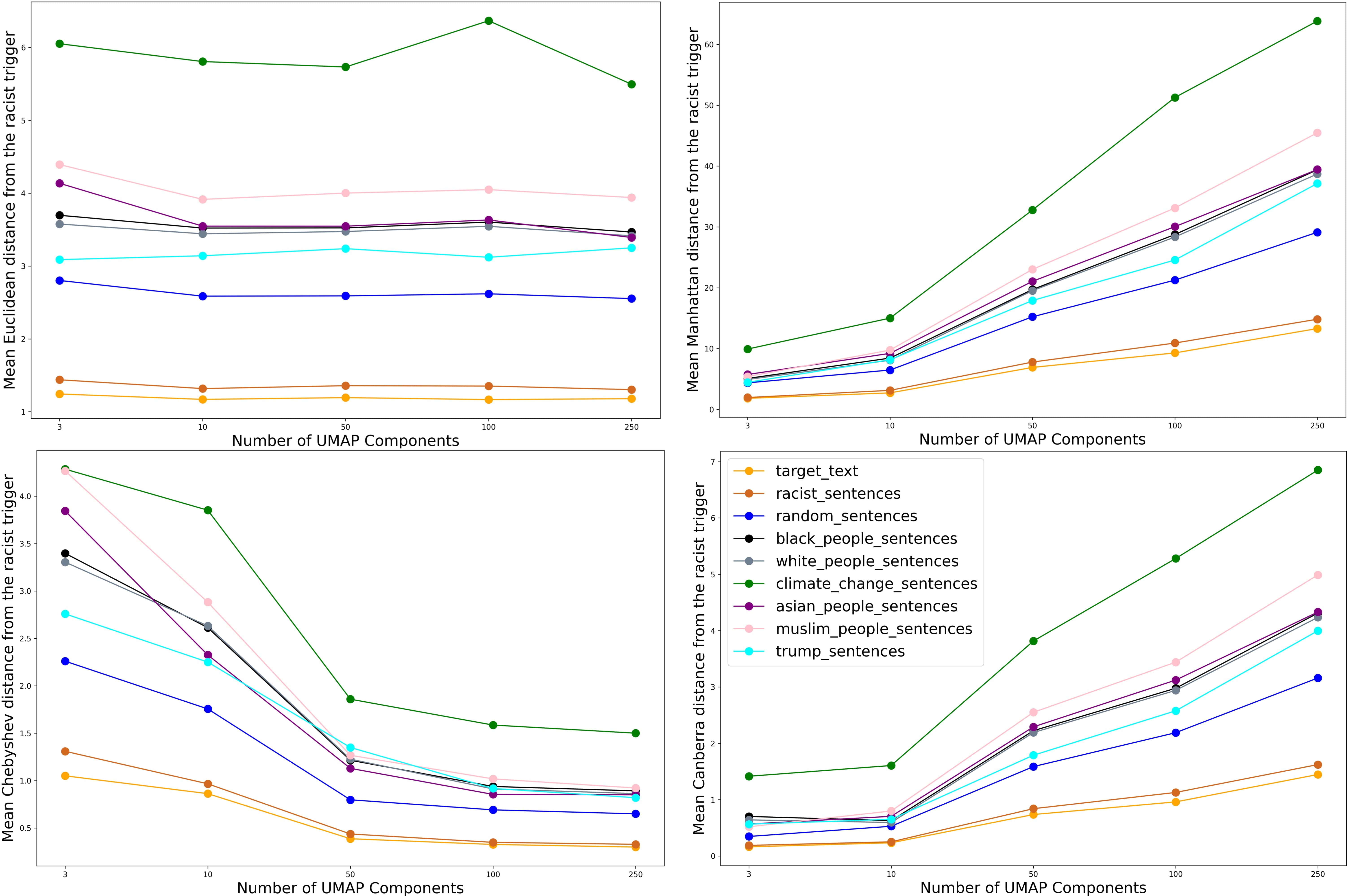}
  \caption{Across a range of UMAP hyperparameters (here, neighbors = 50, minimum distance = 0.25) and a range of reduced UMAP dimensions (3 to 250), we see consistently lower average Euclidean, Manhattan, Chebyshev \& Canberra distances between the trigger, target (orange) and racist (brown) text. Here, average distance is computed as the average of the distances between the trigger and all sentences in a group.}
  \label{fig:sliced-metrics}
\end{figure*}

%%%%%%%%%%%%%%%%%%%%%%%%%%%%%%%%%%%%%%%%% BELOW THIS

\section{Potential Explanations, Limitations \& Future Work}
\label{sec:future-work}

Based on our initial findings supporting the geometric explanation for universal adversarial attacks, we hypothesize potential explanations for observed behavior in these triggers.

\begin{enumerate}
\item We hypothesize that universal adversarial triggers might be effective because the model sees a semantically meaningful adversarial input instead of a gibberish trigger and generates on top of it.
\item Longer triggers were found to be more effective by \citet{Wallace2019}. This could potentially be explained by the fact that more tokens give more room for the trigger to capture semantic information about the adversarial training text.
\item Triggers are seen to transfer across models. We observe that this transferability occurs across a number of models using the same tokenization algorithm. This behavior has also been studied by \citet{zou2023universal} who generate universal adversarial prompts which cause open and proprietary foundation models to generate harmful content. Such an attack has been show to transfer across different model families like ChatGPT, Llama, PaLM, Bard and Claude. Understanding transferability of universal attacks is an open problem and can be an interesting direction for future work.
\end{enumerate}

Despite encouraging initial evidence, we note that experimental techniques involving dimensionality reduction must be considered with caution. The primary limitation of this work is the usage of dimensionality reduction which could be providing an incomplete picture. Future lines of work in interpretability include expanding experimentation to additional techniques. This can include studying the evolution of distances between sentence groups across layers, visualizing the evolution of GPT-2's hidden representations, analyzing the Kullback–Leibler divergence between the distribution of predicted tokens for adversarial and non-adversarial settings, etc.

\section{Conclusion}
In this work, we present a novel geometric perspective which potentially explains gradient-based universal adversarial attacks on large language models. We find initial experimental evidence indicating that these triggers could be behaving like embedding vectors which approximate the semantic information in their adversarial training region. We further leverage this perspective to offer potential explanations for observed behavior of these triggers in literature. With this novel geometric perspective, we aim to shed light on the underlying mechanism driving universal attacks. This would further enable us to understand the failure modes of LLMs and importantly, enable their mitigation.

\section*{Acknowledgements}
We thank the reviewers of `The $2^{\text{nd}}$ New Frontiers in Adversarial Machine Learning Workshop' ICML 2023, for their thoughtful and valuable feedback which helped improve this work. We also thank Yoon Kim, Siddharth Swaroop, Abbas Zeitoun and Ani Nrusimha for their generous technical input and helpful suggestions.

\nocite{langley00}

\bibliography{example_paper}
\bibliographystyle{icml2023}

%%%%%%%%%%%%%%%%%%%%%%%%%%%%%%%%%%%%%%%%%%%%%%%%%%%%%%%%%%%%%%%%%%%%%%%%%%%%%%%
%%%%%%%%%%%%%%%%%%%%%%%%%%%%%%%%%%%%%%%%%%%%%%%%%%%%%%%%%%%%%%%%%%%%%%%%%%%%%%%
% APPENDIX
%%%%%%%%%%%%%%%%%%%%%%%%%%%%%%%%%%%%%%%%%%%%%%%%%%%%%%%%%%%%%%%%%%%%%%%%%%%%%%%
%%%%%%%%%%%%%%%%%%%%%%%%%%%%%%%%%%%%%%%%%%%%%%%%%%%%%%%%%%%%%%%%%%%%%%%%%%%%%%%
\newpage
\appendix
\onecolumn
\section{Appendix}
\label{sec:appendix}

\subsection{Plots for the Racist Trigger}

\noindent\begin{minipage}{\textwidth}
    \centering
    \includegraphics[width=1\textwidth, height=1\textwidth]{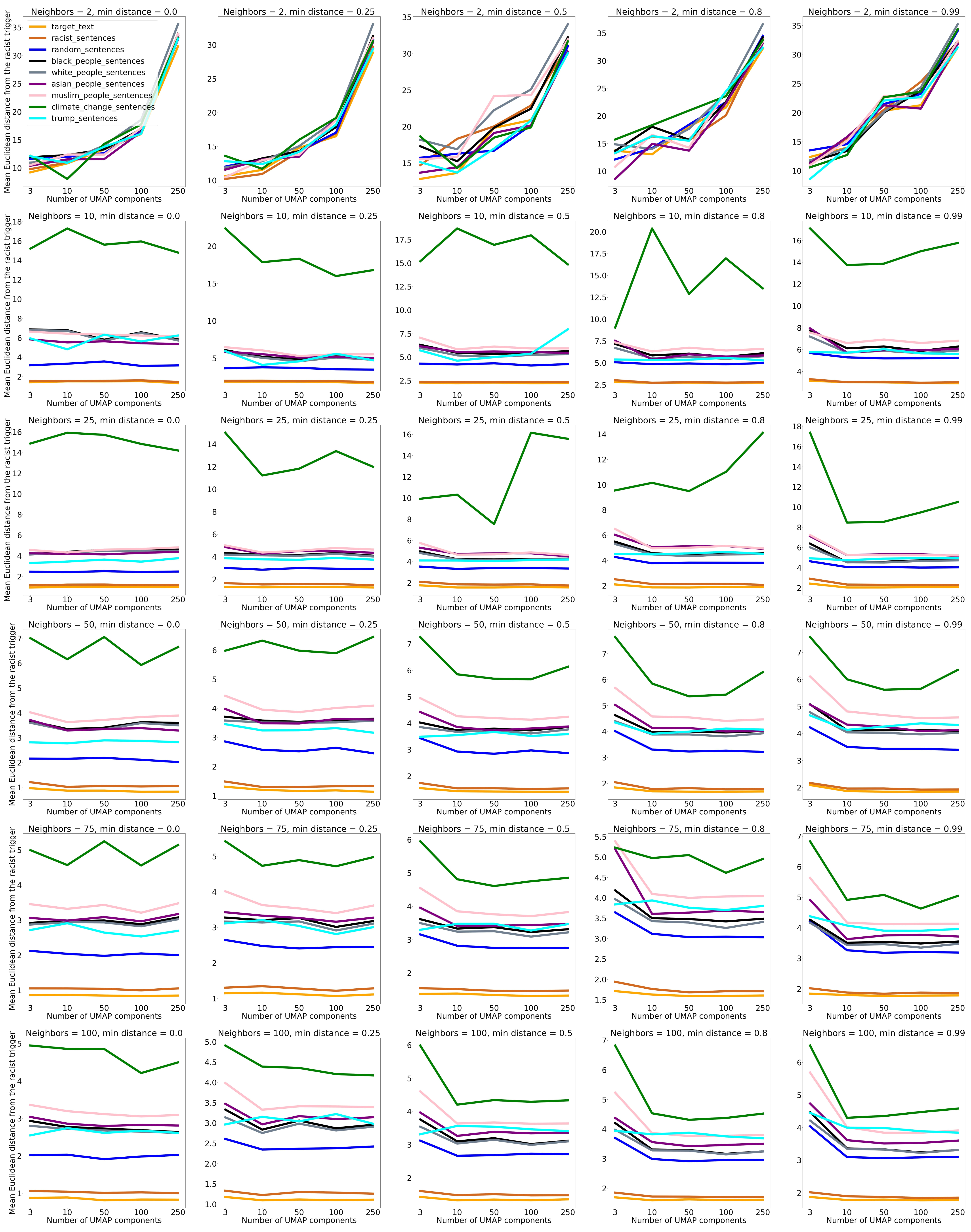}
    \captionof{figure}{Euclidean distance between the racist trigger and other sentence groups for varying UMAP hyperparameters and reduced dimensions.}
\end{minipage}

\begin{figure*}
\begin{center}
\includegraphics[width=1\textwidth, height=1\textwidth]{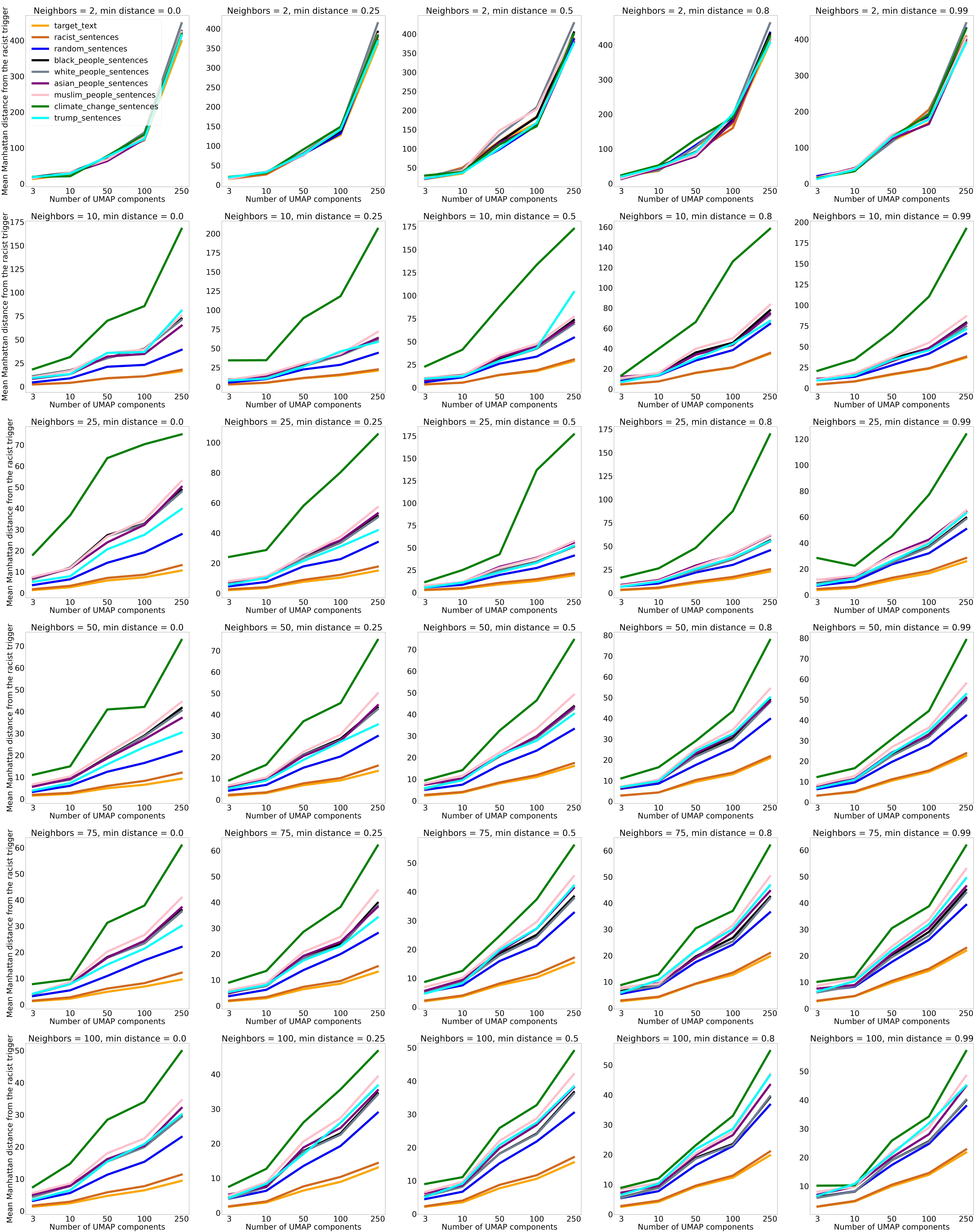}
    \vspace*{-5mm}
    \caption{Manhattan distance between the racist trigger and other sentence groups for varying UMAP hyperparameters and reduced dimensions.} 
    \label{fig:manhattan-racist}
\end{center}
\end{figure*}   

\begin{figure*}
\begin{center}
\includegraphics[width=1\textwidth, height=1\textwidth]{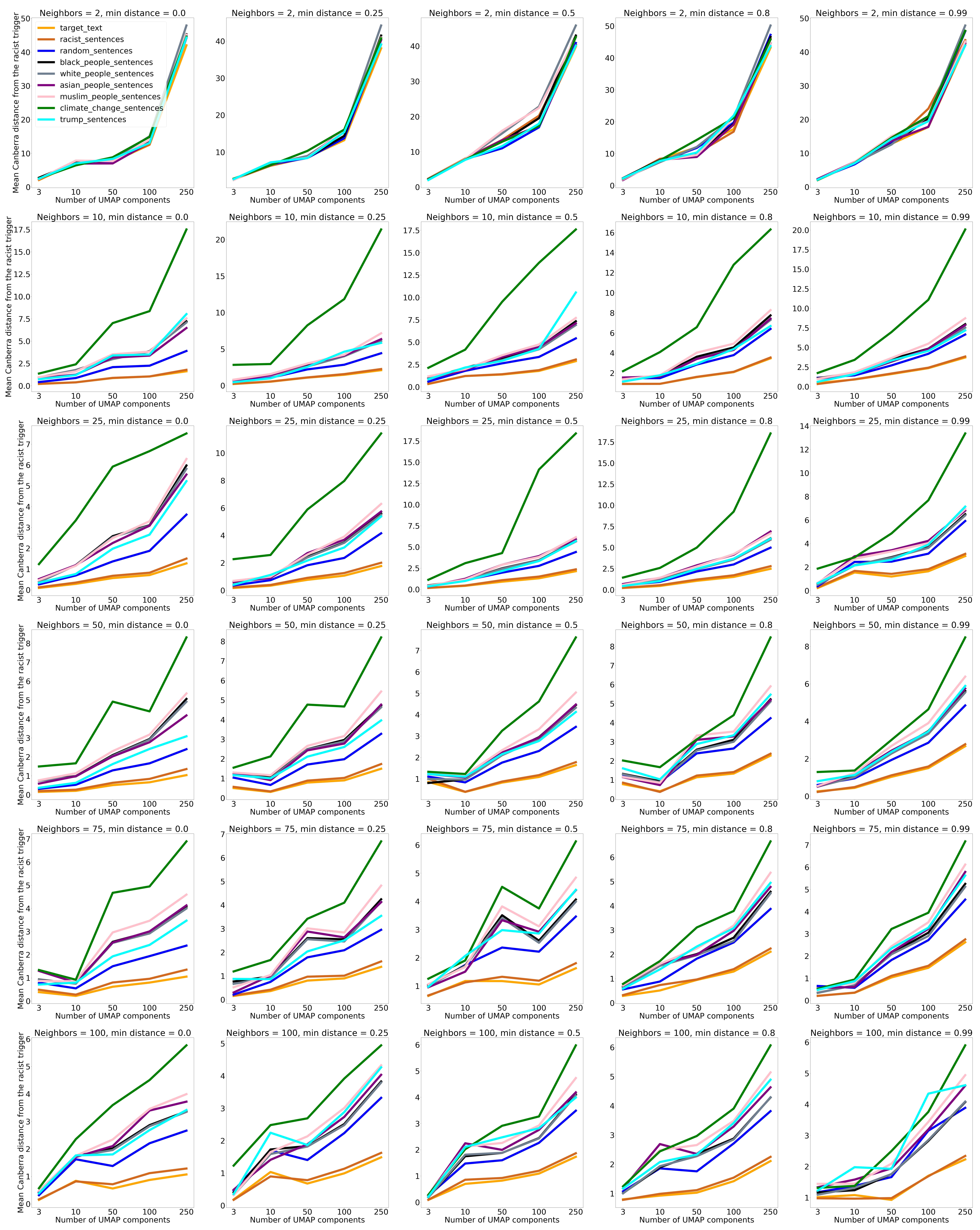}
    \vspace*{-5mm}
    \caption{Canberra distance between the racist trigger and other sentence groups for varying UMAP hyperparameters and reduced dimensions.} 
    \label{fig:canberra-racist}
\end{center}
\end{figure*}

\begin{figure*}
\begin{center}
\includegraphics[width=1\textwidth, height=1\textwidth]{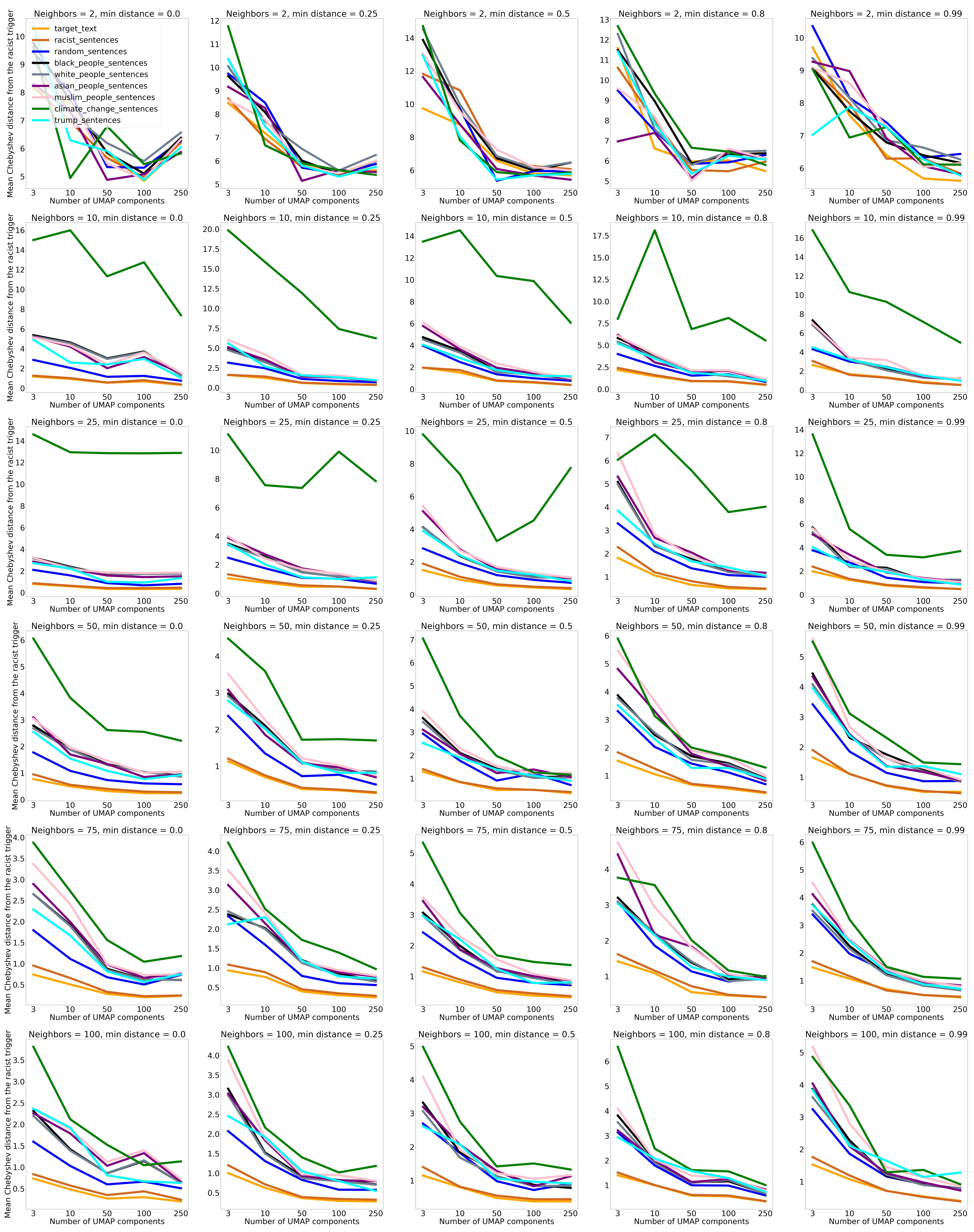}
    \vspace*{-5mm}
    \caption{Chebyshev distance between the racist trigger and other sentence groups for varying UMAP hyperparameters and reduced dimensions.} 
    \label{fig:chebyshev-racist}
\end{center}
\end{figure*}

\newpage

\subsection{Plots for the Flat Earth \& Anti-Vaccination Triggers}

\begin{figure*}[!htb]
\begin{center}
\includegraphics[width=1\textwidth, height=1\textwidth]{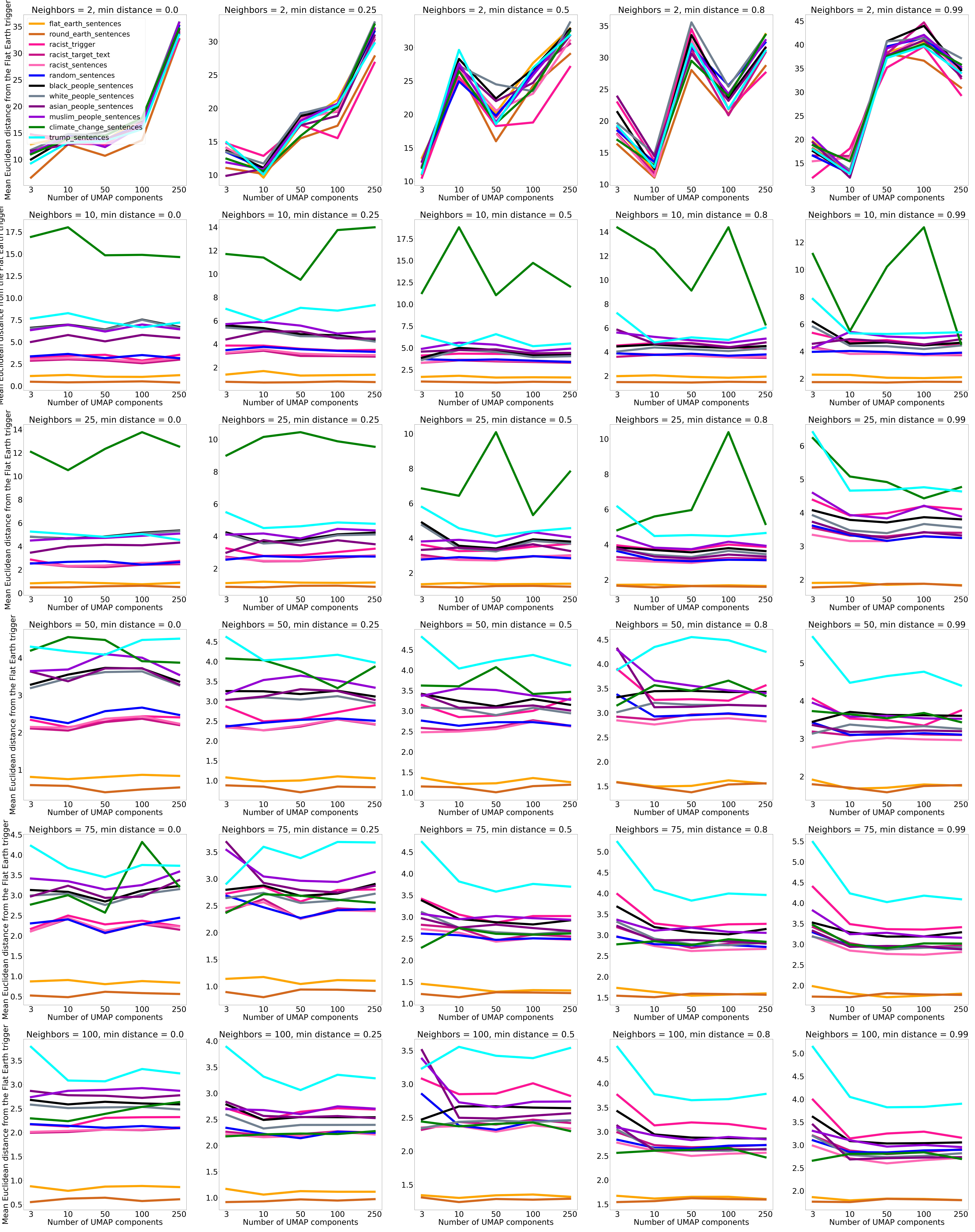}
    \vspace*{-5mm}
    \caption{Euclidean distance between the Flat Earth trigger and other sentence groups for varying UMAP hyperparameters and reduced dimensions.} 
    \label{fig:euclidean-flatearth}
\end{center}
\end{figure*} 

\begin{figure*}[!htb]
\begin{center}
\includegraphics[width=1\textwidth, height=1\textwidth]{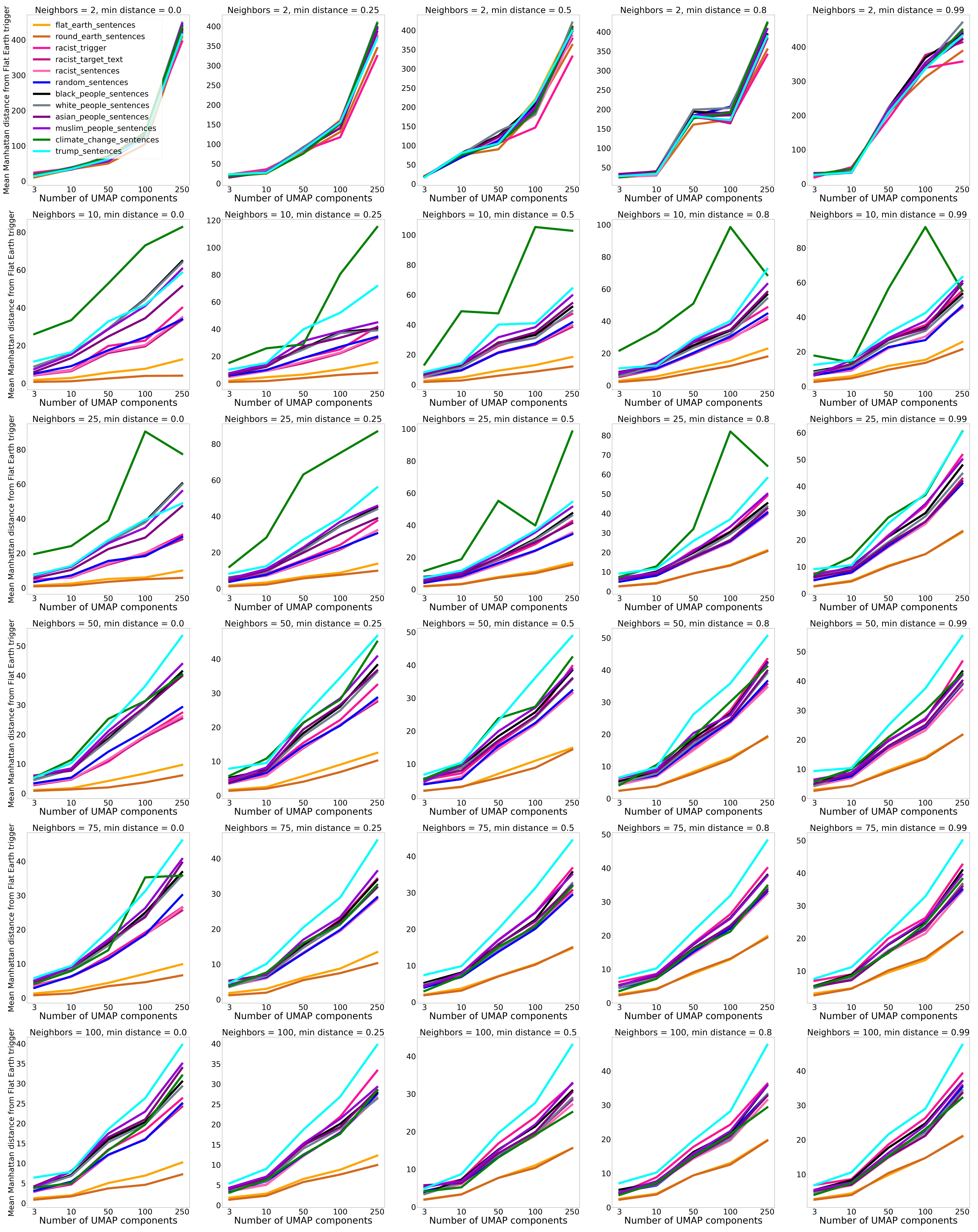}
    \vspace*{-5mm}
    \caption{Manhattan distance between the Flat Earth trigger and other sentence groups for varying UMAP hyperparameters and reduced dimensions.} 
    \label{fig:manhattan-flatearth}
\end{center}
\end{figure*}   

\begin{figure*}[!htb]
\begin{center}
\includegraphics[width=1\textwidth, height=1\textwidth]{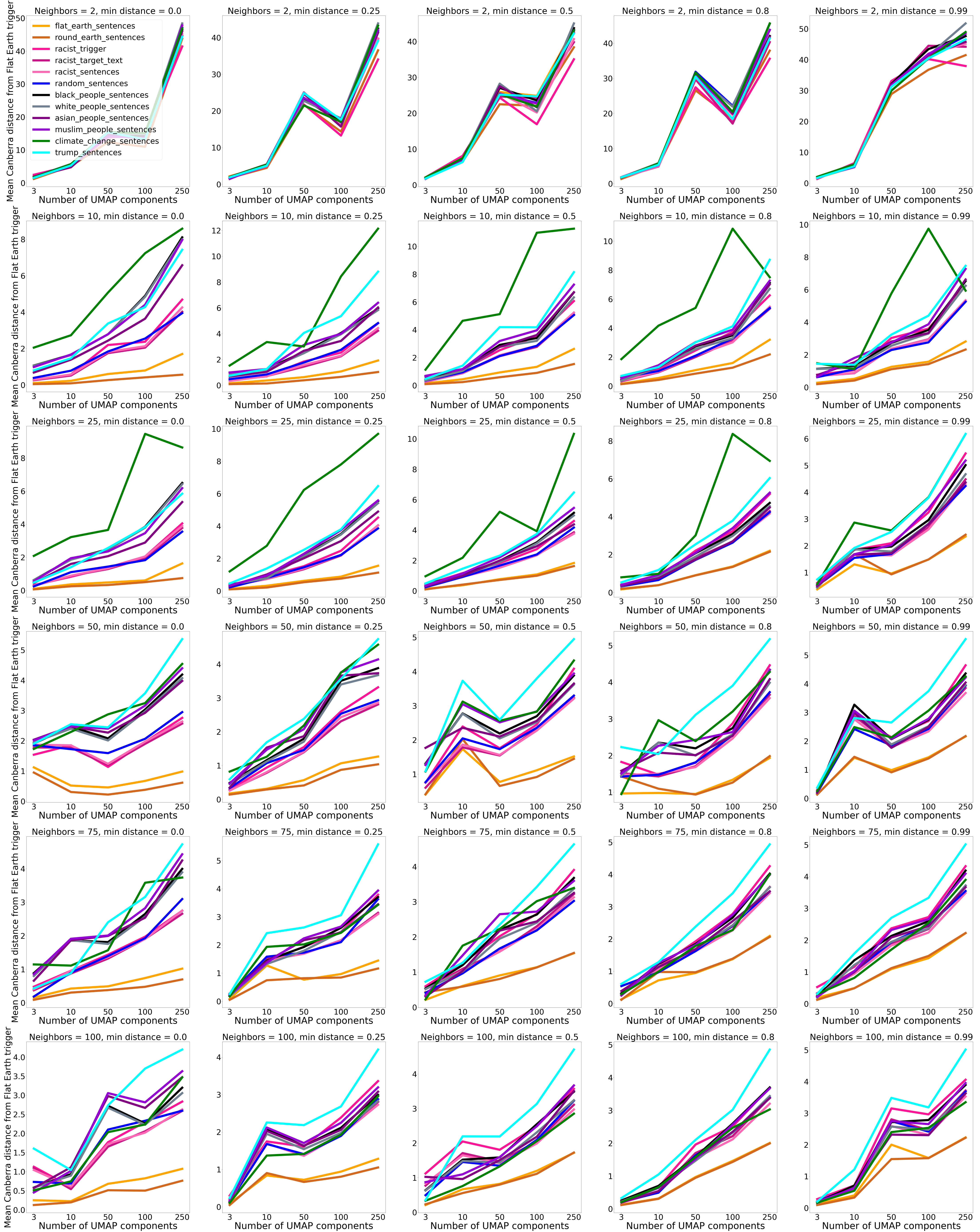}
    \vspace*{-5mm}
    \caption{Canberra distance between the Flat Earth trigger and other sentence groups for varying UMAP hyperparameters and reduced dimensions.} 
    \label{fig:canberra-flatearth}
\end{center}
\end{figure*}

\begin{figure*}[!htb]
\begin{center}
\includegraphics[width=1\textwidth, height=1\textwidth]{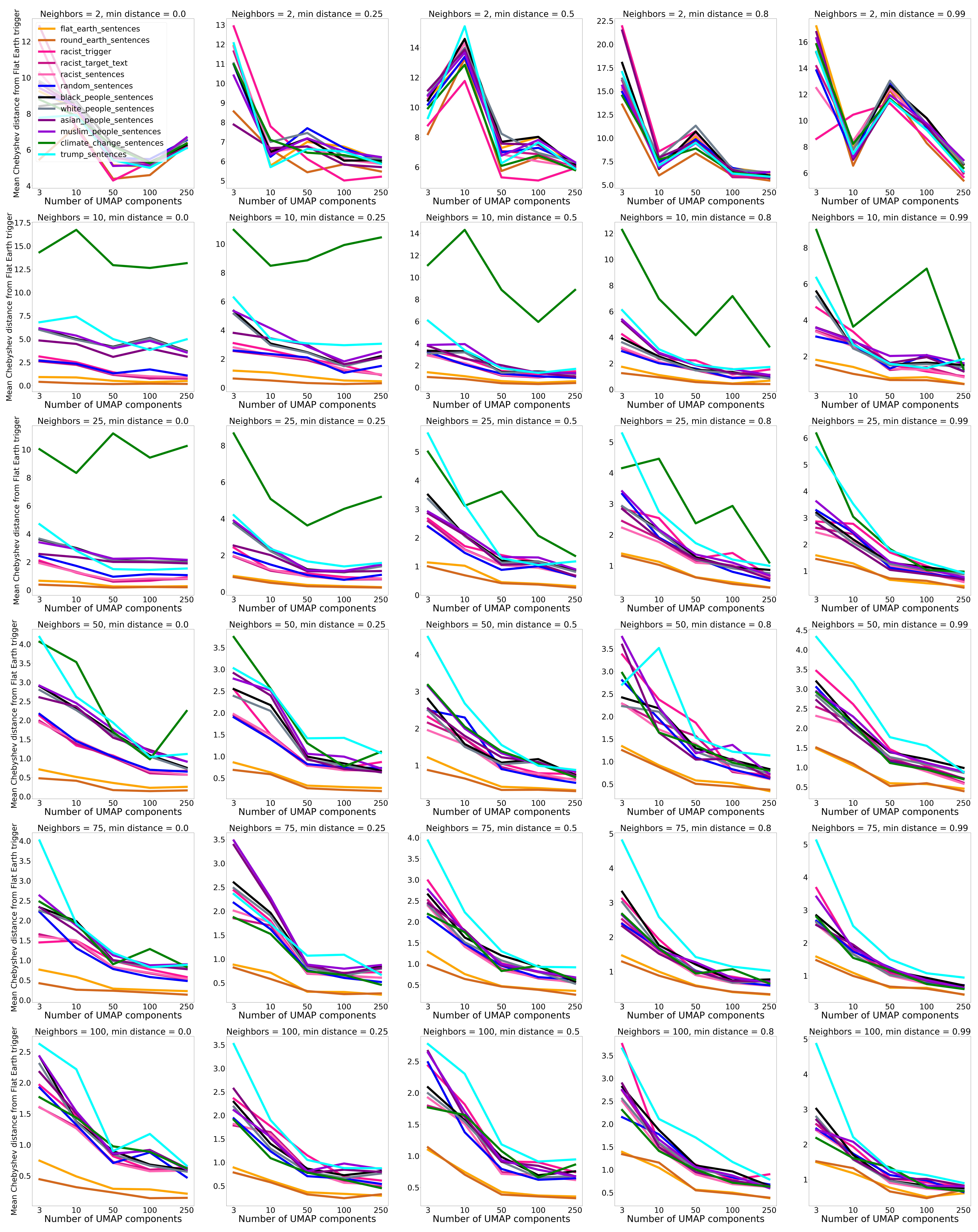}
    \vspace*{-5mm}
    \caption{Chebyshev distance between the Flat Earth trigger and other sentence groups for varying UMAP hyperparameters and reduced dimensions.} 
    \label{fig:chebyshev-flatearth}
\end{center}
\end{figure*}

\begin{figure*}[!htb]
\begin{center}
\includegraphics[width=1\textwidth, height=1\textwidth]{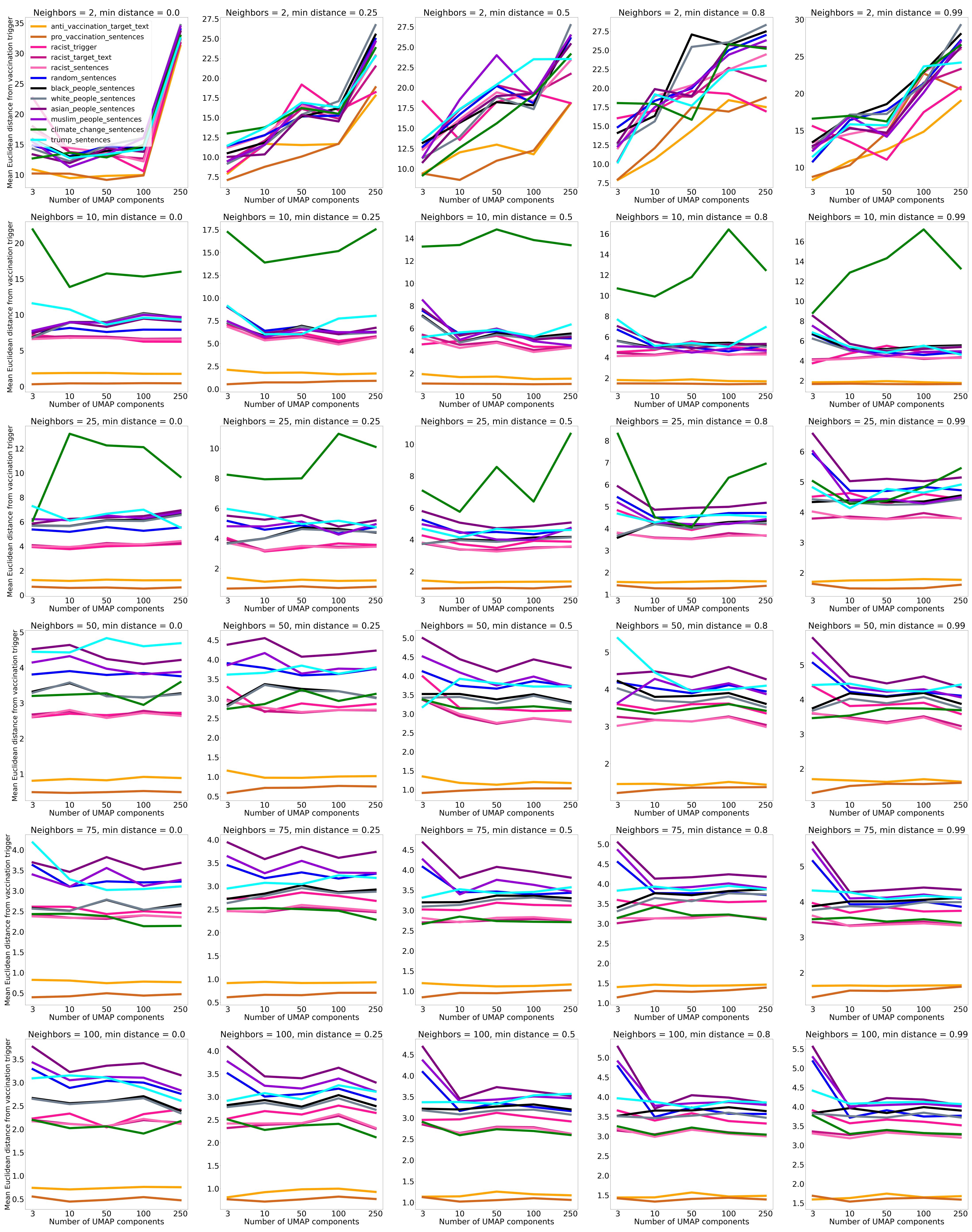}
    \vspace*{-5mm}
    \caption{Euclidean distance between the vaccination trigger and other sentence groups for varying UMAP hyperparameters and reduced dimensions.} 
    \label{fig:euclidean-vaccination}
\end{center}
\end{figure*} 

\begin{figure*}[!htb]
\begin{center}
\includegraphics[width=1\textwidth, height=1\textwidth]{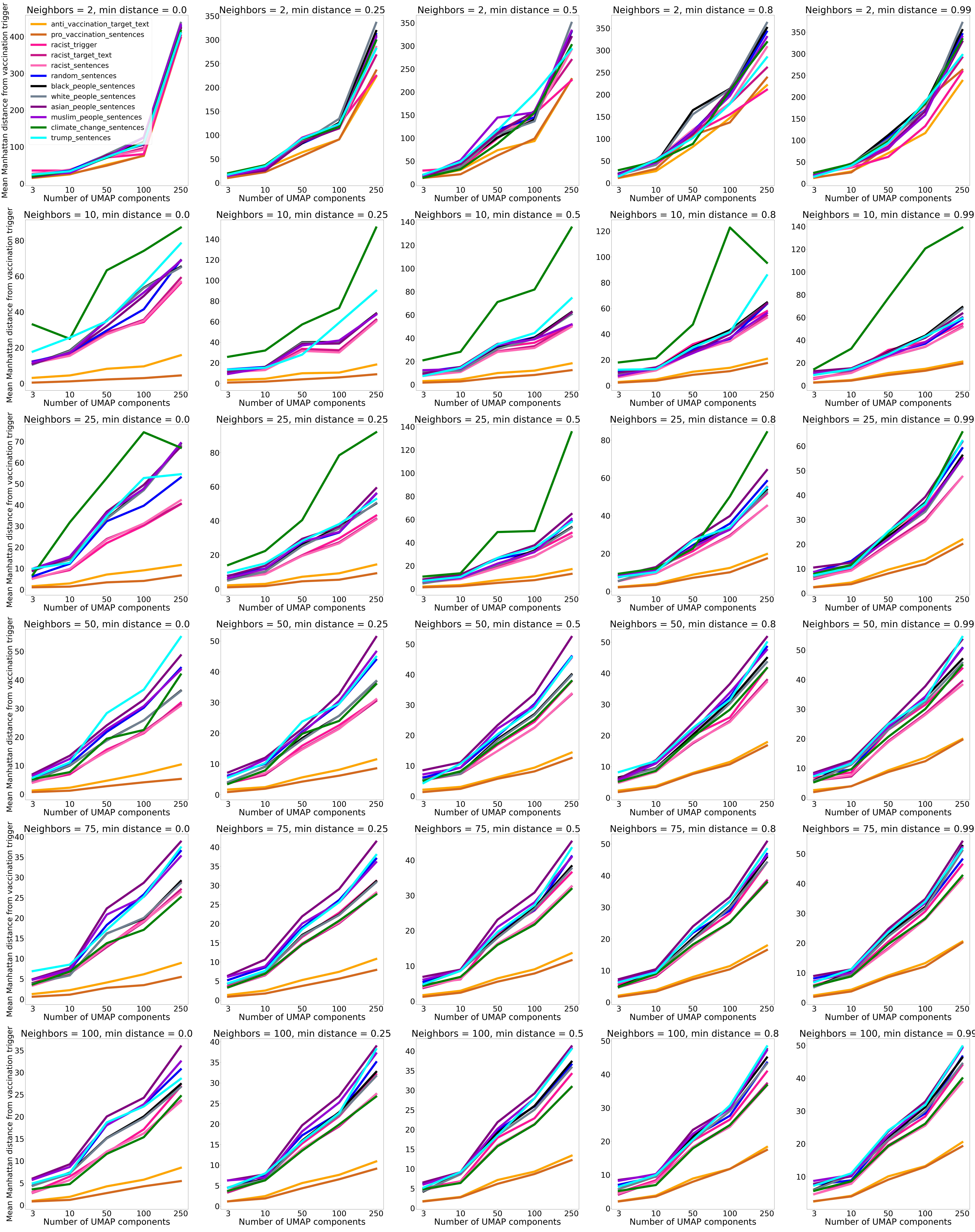}
    \vspace*{-5mm}
    \caption{Manhattan distance between the vaccination trigger and other sentence groups for varying UMAP hyperparameters and reduced dimensions.}
    \label{fig:manhattan-vaccination}
\end{center}
\end{figure*}

\begin{figure*}[!htb]
\begin{center}
\includegraphics[width=1\textwidth, height=1\textwidth]{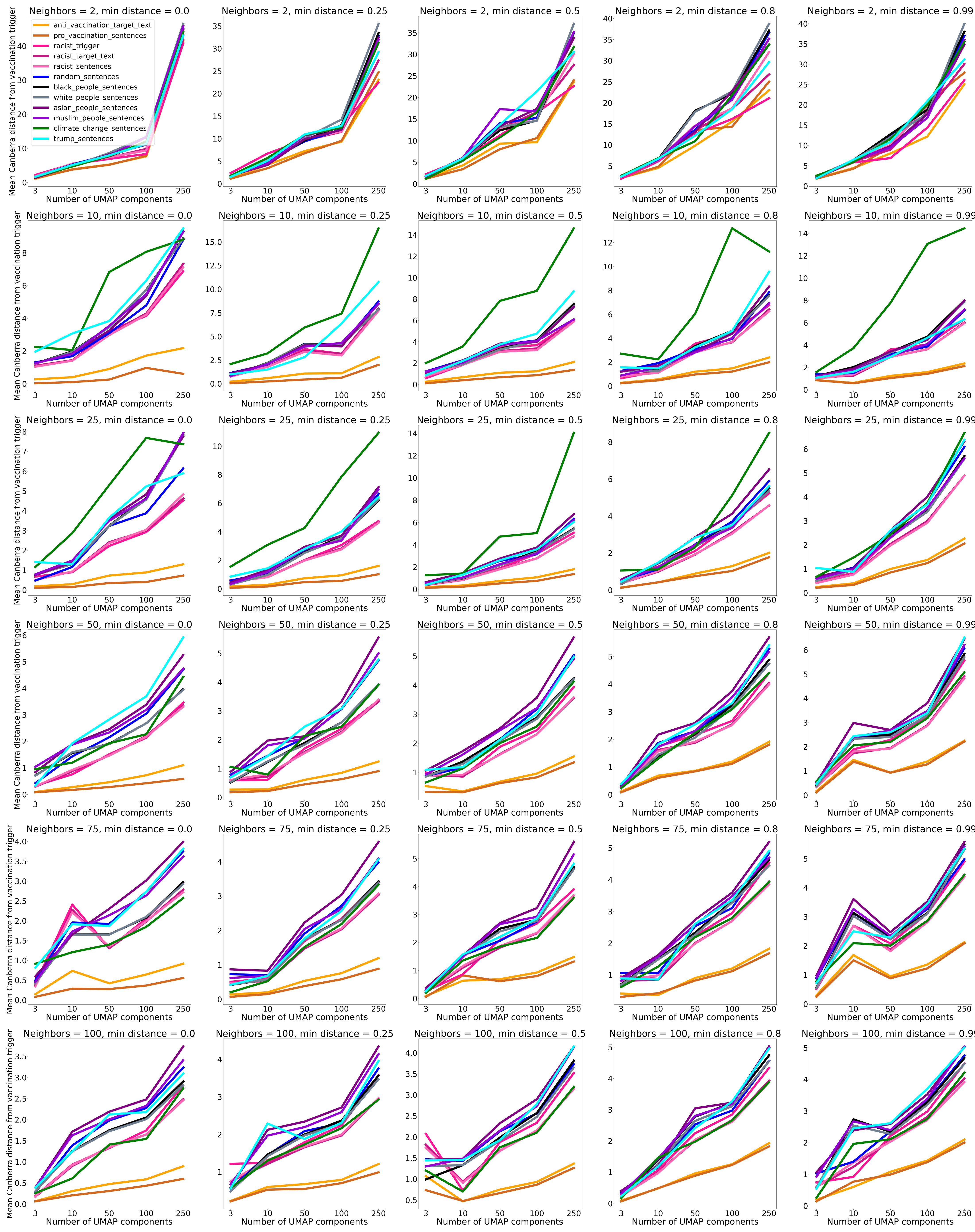}
    \vspace*{-5mm}
    \caption{Canberra distance between the vaccination trigger and other sentence groups for varying UMAP hyperparameters and reduced dimensions.}
    \label{fig:canberra-vaccination}
\end{center}
\end{figure*}

\begin{figure*}[!htb]
\begin{center}
\includegraphics[width=1\textwidth, height=1\textwidth]{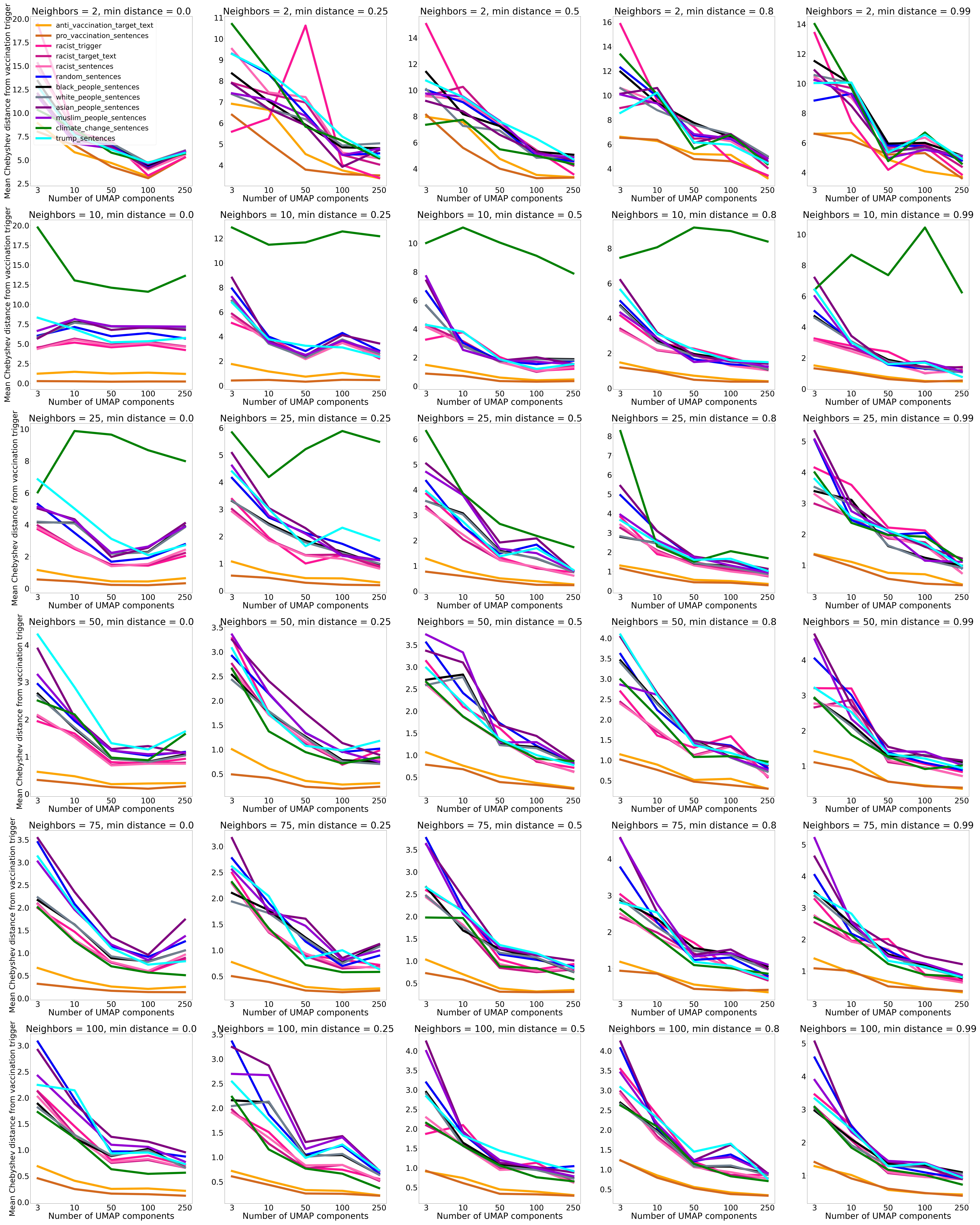}
    \vspace*{-5mm}
    \caption{Chebyshev distance between the vaccination trigger and other sentence groups for varying UMAP hyperparameters and reduced dimensions.} 
    \label{fig:chebyshev-vaccination}
\end{center}
\end{figure*}

%%%%%%%%%%%%%%%%%%%%%%%%%%%%%%%%%%%%%%%%%%%%%%%%%%%%%%%%%%%%%%%%%%%%%%%%%%%%%%%
%%%%%%%%%%%%%%%%%%%%%%%%%%%%%%%%%%%%%%%%%%%%%%%%%%%%%%%%%%%%%%%%%%%%%%%%%%%%%%%

\end{document}